\pdfoutput=1

\documentclass[11pt]{article}

\usepackage[review]{EMNLP2022}
\usepackage{amsmath}
\usepackage{times}
\usepackage{latexsym}
\usepackage{color}
\usepackage{arydshln}
\usepackage{bbding}
\usepackage{amssymb}
\usepackage{bigstrut}
\usepackage{stfloats}
\usepackage{inconsolata}
\usepackage{arydshln}
\usepackage{bbding}
\usepackage{svg}
\usepackage[T1]{fontenc}
\usepackage{bm}
\usepackage{enumitem}
\usepackage{blindtext}
\usepackage[utf8]{inputenc}
\usepackage{url}
\usepackage{mdwlist}
\usepackage{algorithm}
\usepackage{algpseudocode}
\usepackage{amsmath}
\usepackage{microtype}
\usepackage{multirow}
\usepackage{CJK}
\usepackage{setspace}
\usepackage{graphicx}  
\usepackage{float}  
\usepackage{subfigure} 
\usepackage{booktabs}

\title{Improving Semantic Matching  through Dependency-Enhanced \\Pre-trained Model with Adaptive Fusion}

 \author{Jian Song\textsuperscript{1}\thanks{\ \ Equal contribution.} \and Di Liang\textsuperscript{2}\footnotemark[1] \and Rumei Li\textsuperscript{2} \and Yuntao Li\textsuperscript{2} \and Sirui Wang\textsuperscript{2} \\
{\bf Minlong Peng\textsuperscript{3} \and Wei Wu\textsuperscript{2} \and Yongxin Yu\textsuperscript{1}\thanks{ \ \ Corresponding author.}} \\
$^1$Tianjin University, Tianjin, China \quad $^2$Meituan Inc., Beijing, China \\ $^3$Fudan University, Shanghai, China \\ \{songjian799, yyx\}@tju.edu.cn \\
\{liangdi04, lirumei, wangsirui, liyuntao, wuwei30\}@meituan.com \\
\{mlpeng16\}@fudan.edu.cn 
}

\begin{document}

\maketitle

\begin{abstract}

Transformer-based pre-trained models like BERT have achieved great progress on Semantic Sentence Matching. Meanwhile, dependency prior knowledge has also shown general benefits in multiple NLP tasks. However, how to efficiently integrate dependency prior structure into pre-trained models to better model complex semantic matching relations is still unsettled.
In this paper, we propose the \textbf{D}ependency-Enhanced \textbf{A}daptive \textbf{F}usion \textbf{A}ttention (\textbf{DAFA}), which explicitly introduces dependency structure into pre-trained models and adaptively fuses it with semantic information.
Specifically, \textbf{\emph{(i)}} DAFA first proposes a structure-sensitive paradigm to construct a dependency matrix for calibrating attention weights.
It adopts an adaptive fusion module to integrate the obtained dependency information and the original semantic signals. 
Moreover, DAFA reconstructs the attention calculation flow and provides better interpretability. 
By applying it on BERT, our method achieves state-of-the-art or competitive performance on 10 public datasets, demonstrating the benefits of adaptively fusing dependency structure in semantic matching task.
\end{abstract}

\section{Introduction}
Semantic Sentence Matching (SSM) is a fundamental technology in multiple NLP scenarios. The goal of SSM is to compare two sentences and identify their semantic relationship. 
It is widely utilized in recommendation systems \cite{zeng2021zero}, dialogue systems \cite{yu2014cognitive}, search systems \cite{li2014semantic}, and so on \cite{gao2018neural}. 

\begin{figure}
\centering
\includegraphics[width=0.46\textwidth]{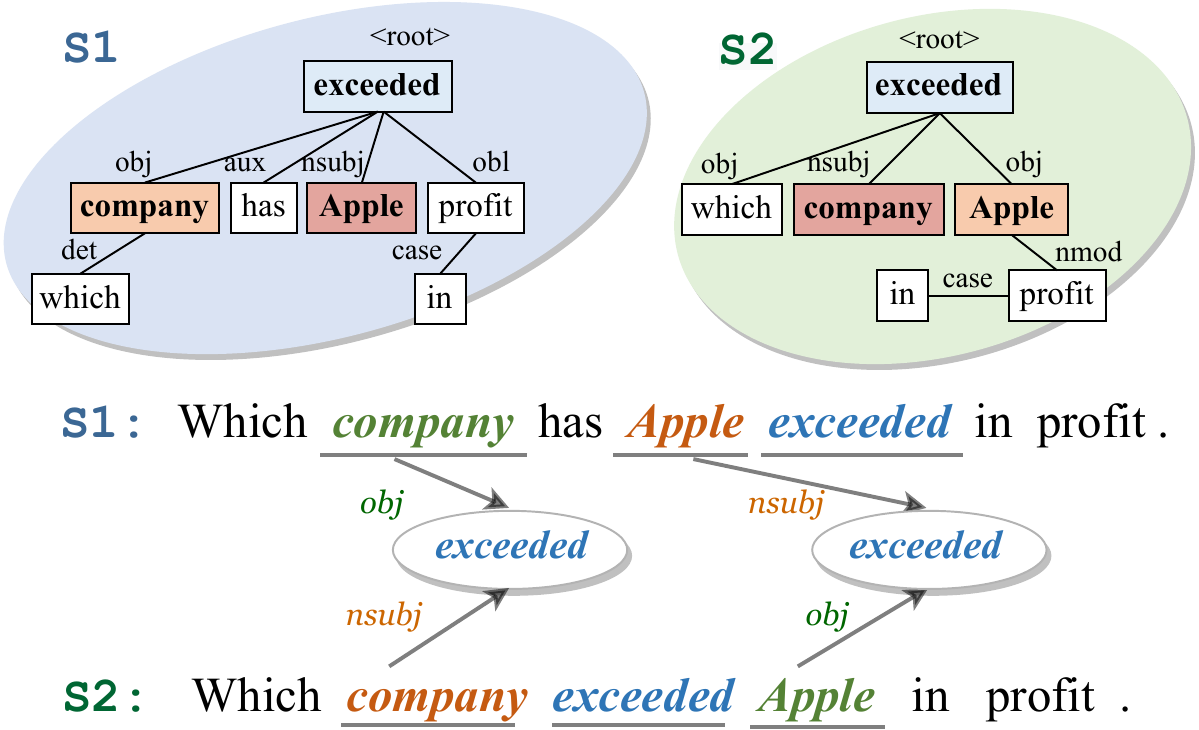}
\caption{\label{fig:example} Example sentences that have similar literals, but express different semantics caused by inconsistent dependency.}
\vspace{-0.54cm}
\end{figure}

Across the rich history of semantic matching research, there have been two main streams of studies for solving this problem. One is to utilize a sentence encoder to convert sentences into low-dimensional vectors in the latent space, and apply a parameterized function to learn the matching scores between them \cite{conneau2017supervised,reimers2019sentence}. Another paradigm tends to align phrases and aggregate the integrated information into prediction layer to acquire similarity and make a sentence-level decision \cite{chen2016enhanced,tay2017compare}. 
After the emergence of large-scale pre-trained language models (PLMs), recent work attempts to integrate external knowledge \cite{miller1995wordnet,bodenreider2004unified} into PLMs. For example, SemBERT \cite{zhang2020semantics} concatenates semantic role annotation to enhance BERT. UERBERT \cite{xia2021using} chooses to inject synonym knowledge. SyntaxBERT \cite{bai2021syntax} integrates the syntax tree into transformer-based models. Meanwhile, leveraging external knowledge to enhance PLMs has been proven to be highly useful for multiple NLP tasks \cite{kiperwasser2018scheduled,bowman2016fast}.

Although previous work has achieved great progress in SSM, existing models (e.g., \textit{BERT}, \textit{RoBERTa}) still cannot efficiently and explicitly utilize dependency structure to identify semantic differences, especially when two sentences are literally similar. To illustrate that, we display an example of misjudgment by BERT \cite{devlin2018bert} in Figure \ref{fig:example}. 
In the first sentence, the dependency between \emph{“exceeded”} and \emph{“company”} is \textsf{\texttt{obj}}, between \emph{“exceeded”} and \emph{“Apple”} is \textsf{\texttt{nsubj}}. Its dependency structure is completely opposite to the second sentence. 
Although the literal similarity of these two sentences is extremely high, the semantics are still quite different. 
Based on the above observations, we intuitively believe that the dependency structure needs to be considered in the process of semantic matching. From a deeper perspective, the MLM training approach of most existing PLMs is optimizing the co-occurrence probability statistically \cite{ yang2017breaking}, but dependency structure can reflect the dependency relationship within the sentence and integrate prior language knowledge to enhance interaction features.
Combined with the actual attention alignment process, we believe that constructing a dependency matrix, \textsl{\textbf{\textit{strengthening the attention weight of same dependency and reducing the attention weight of different dependency}}}, will further improve the performance of existing PLMs. Therefore, two systemic questions arise naturally:

\textsl{\textbf{\textit{Q1:} {How to construct a dependency matrix that contains dependency prior knowledge?}}}
Inconsistent dependency structures can lead to severe semantic divergence even between sentences with similar text.
To capture the dependency structure features, we propose a structure-aware paradigm to construct the dependency matrix. Our paradigm utilizes three distinct modules, including the dependency similarity between words, the matching of dependency subgraphs, and the \textit{tf-idf} weights.

\textsl{\textbf{\textit{Q2:} {How to integrate the introduced dependency signals provided by dependency matrix?}}}
To maximize the benefits of the dependency knowledge, we integrate the dependency structure to calibrate our attention alignment. Therefore, our model can not only understand sentence semantics, but also further enhance structural alignment awareness.
However, a hard fusion of dependency and semantic signals by simple structure may break the original representing ability of PLMs. How to inject the obtained dependency information softly remains a hard issue. In this paper, we propose an Adaptive Fusion module: \textbf{\textit{(i)}} It first inter-aligns these two signals through distinct attentions, and generates vectors describing sentence matching details. \textbf{\textit{(ii)}} Then, multiple gates are utilized to extract meaningful information adaptively. \textbf{\textit{(iii)}} Moreover, our vectors are further scaled with another fuse-gate to reduce the possibility of noise introduced by dependency features. Eventually, this soft aggregation method can adaptively fuse these collected information and obtain the fusion signals.

\begin{figure*}
\centering
\includegraphics[width=0.97\textwidth]{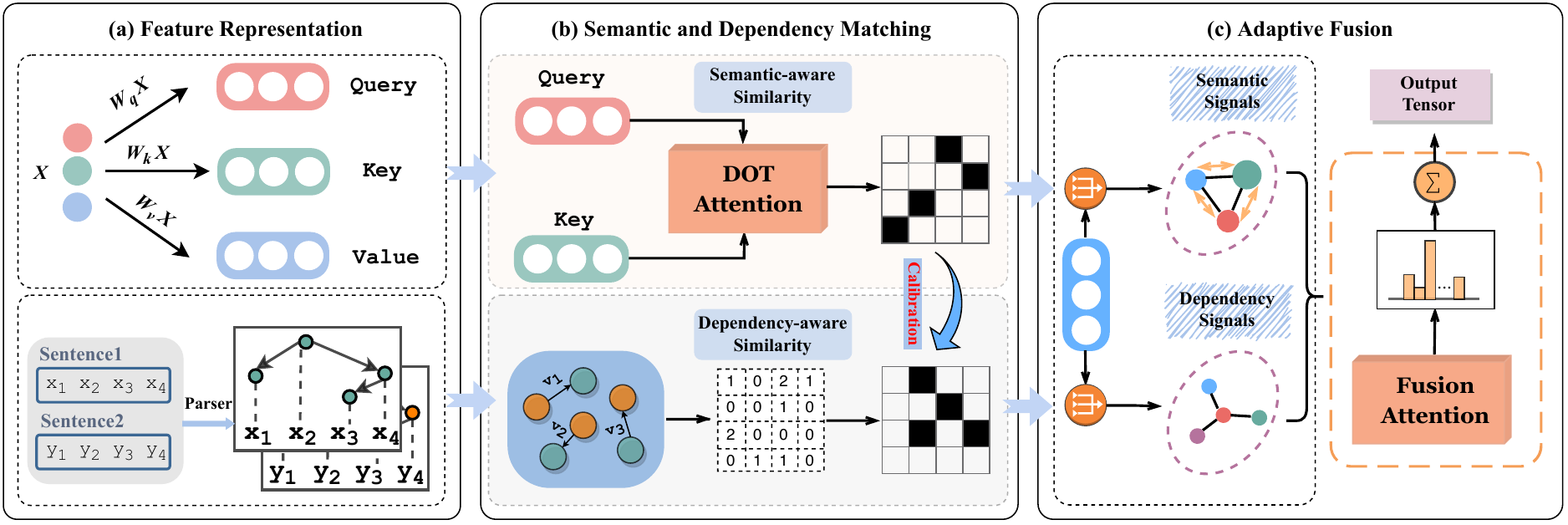}\caption{\label{fig:Overview}The overall architecture of Dependency-enhanced Adaptive Fusion Attention. It is the structure after we apply DAFA to the original multi-head attention.}
\vspace{-0.45cm}
\end{figure*}

Overall, our contributions are mainly as follows:
\begin{itemize}[itemsep= 0 pt,topsep = 3 pt]
\setlength{\parskip}{0pt}
\item We discuss in detail the methodology of further leveraging dependency and explicitly integrating dependency structure into PLMs. 
\item We propose a novel dependency calibration and fusion network DAFA, which is a meaningful practice combining semantic and dependency information and provides better interpretability. DAFA leverages dependency structure to calibrate attention alignment and constructs a fusion module to adaptively integrate distinct features.
\item To verify the effectiveness of DAFA, we conduct extensive experiments on 10 datasets in SSM and achieves state-of-the-art or competitive performance over other strong baselines. It proves the effectiveness of our method.
\end{itemize}

\section{Approach}
In this section, we introduce DAFA in detail and the overall architecture is presented in Figure \ref{fig:Overview}.

\subsection{How to Build the Dependency Matrix}
To construct a dependency matrix that contains dependency knowledge, we propose a structure-aware paradigm with three modules: \textit{(1)} we first use the similarity of dependency trees to build our matrix. \textit{(2)} Then, we introduce subgraph matching to further align the dependency substructures. \textit{(3)} Moreover, we also add \textit{tf-idf} weights to reallocate more attention to keywords and their dependency. 

We utilize trigrams to model dependency trees. One trigram unit denotes one branch.
In the first sentence of Figure \ref{fig:example}, \{\emph{“exceeded”}, \textsf{\texttt{nsubj}}, \emph{“Apple”}\} is a trigram unit. Apart from literal similarity, two similar trigrams indicate closer semantics. We set \textit{n} and \textit{m} as the lengths of sentence $A$ and $B$. $A^i$ denotes the \textit{i}-th word of $A$ and $\mathcal{D}_A^i$ denotes the trigram with $A^i$ as the tail node. \textit{$h_A^i$}, \textit{$t_A^i$}, and \textit{$r_A^i$} are indicate the head, tail and type node of $\mathcal{D}_A^i$ respectively. We first utilize the dependency trees similarity to build our matrix $\mathbf{M} \in \mathbb{R}^{{n*m}}$:

{\vspace{-0.3cm} 
\setlength{\abovedisplayskip}{0.1cm}
\setlength{\belowdisplayskip}{0.2cm}
\begin{gather}
\hspace{-0.2cm}
\!\!\!
\mathbf{M}(i, j)=
\underbrace{[s(h^i_A, h^j_B) + s(t^i_A, t^j_B)]}_{\textit{head and tail node}} * 
\underbrace{r(r^i_A, r^j_B)}_{\textit{type node}}, \!\!\! \\
s(a, b) = \{1\ or\ 0\, |\, if\ a = b\ or\ \underline{\textup{otherwise}} \}, \\
r(a, b) = \{\theta\ or\ 1\, |\, if\ a = b\ or\ \underline{\textup{otherwise}} \},
\end{gather}}where $s$, $r$ are binary functions and $\theta$ is a constant that determines the effect of dependency type match. 
However, $\mathbf{M}$ may over-focus on dependency match and neglect the comparison of consecutive dependency trigrams. Therefore, by adopting the subgraph matching mechanism, we can also align the substructures of two dependency trees to acquire the continuous dependency similarity $\mathbf{S}$:

{\vspace{-0.3cm} 
\setlength{\abovedisplayskip}{0.1cm}
\setlength{\belowdisplayskip}{0.2cm}
\begin{equation}
  \begin{aligned}
  \mathbf{S}(A^i, B^j) =& \{ \alpha \underline{s(A^i, B^j)}+\nu \underline{\mathbf{S}(A^i_x, B^j_y)} \, |\, \\
  &\forall A^i_x \in \mathcal{T}_A^i \  \textup{and} \  B^j_y \in \mathcal{T}_B^j, \\ 
  &if \ t_A^i=t_B^j \  \textup{and} \  r_A^i = r_B^j \}
  \end{aligned}
\end{equation}}$\nu$ is a decay factor in case $\mathbf{S}$ extremely increase and $\alpha$ is a fixed score to the child nodes of matching trigram pair. $\mathcal{T}_A^i$ denotes the set of child nodes of $A^i$ and $A^i_x$ is the \textit{x}-th child. $\mathbf{S}$ recursively compares the child nodes of matching pair.

However, our dependency matrix still ignores the difference in importance between words in same sentence. Keywords and their dependency relationship should be allocated more attention. Therefore, we rely on the \textit{tf-idf} weights \cite{ramos2003using} to obtain the informative scores of distinct sentence components and align the \textit{tf-idf} weights of tail nodes in two trigrams:

{\vspace{-0.2cm} 
\setlength{\abovedisplayskip}{0.1cm}
\setlength{\belowdisplayskip}{0.1cm}
\begin{equation}
 \begin{aligned}
\mathbf{M_F}(i, j)= \underbrace{\, |\,\mathbf{M}(i, j)+\mathbf{S}(A^i, B^j)\, |\,}_{\textit{dependency and subgraph}}*\underbrace{(tf_{A^i}*tf_{B^j})}_{\textit{tf-idf weights}} 
 \end{aligned}
\end{equation}}where $tf_{a}$ denotes the \textit{tf-idf} weight of $a$. 
And $\mathbf{M_F} \in \mathbb{R}^{{n*m}}$ is our final dependency matrix.

\subsection{How to Integrate Dependency Information}

To better utilize the gained dependency information, we propose to inject our dependency matrix into the original transformer attention module and apply the dependency structure to calibrate the attention alignment. Attention module can be considered as a mapping from query vector $\mathbf{\mathcal{Q}}$ and a set of key-value vector pairs ($\mathbf{\mathcal{K}}$, $\mathbf{\mathcal{V}}$) to the attention distribution. Each layer has multiple parallel attention heads. By introducing $\mathbf{M_F}$, the calculation flow of each head is as follows:

{\vspace{-0.4cm} 
\setlength{\abovedisplayskip}{0.1cm}
\setlength{\belowdisplayskip}{0.2cm}
\begin{equation}
\begin{aligned}
\mathbf{Sem} &= softmax(\frac{\mathbf{\mathcal{Q K}}^T}{\sqrt{d_k}}) * \mathbf{\mathcal{V}}, \\
\mathbf{Dep} &= softmax(\frac{\mathbf{\mathcal{Q K}}^T \odot \mathbf{M_F}}{\sqrt{d_k}}) * \mathbf{\mathcal{V}},
 \end{aligned}
\end{equation}}where $d_k$, $d_v$ is the dimension of $\mathbf{\mathcal{K}}$, $\mathbf{\mathcal{V}}$ and $d_{seq}$ is the input length. We change the dimension of $\mathbf{M_F}$ by adding 1, and ensure each element is in the corresponding position in the sentence alignment.
$\odot$ is the element-wise dot product, and $\mathbf{Dep}\in\mathbb{R}^{{d_v}*d_{seq}}$ denotes the dependency signals from the attention matrix calibrated by our dependency matrix. 

\begin{figure}
\centering
\includegraphics[width=0.30\textwidth]{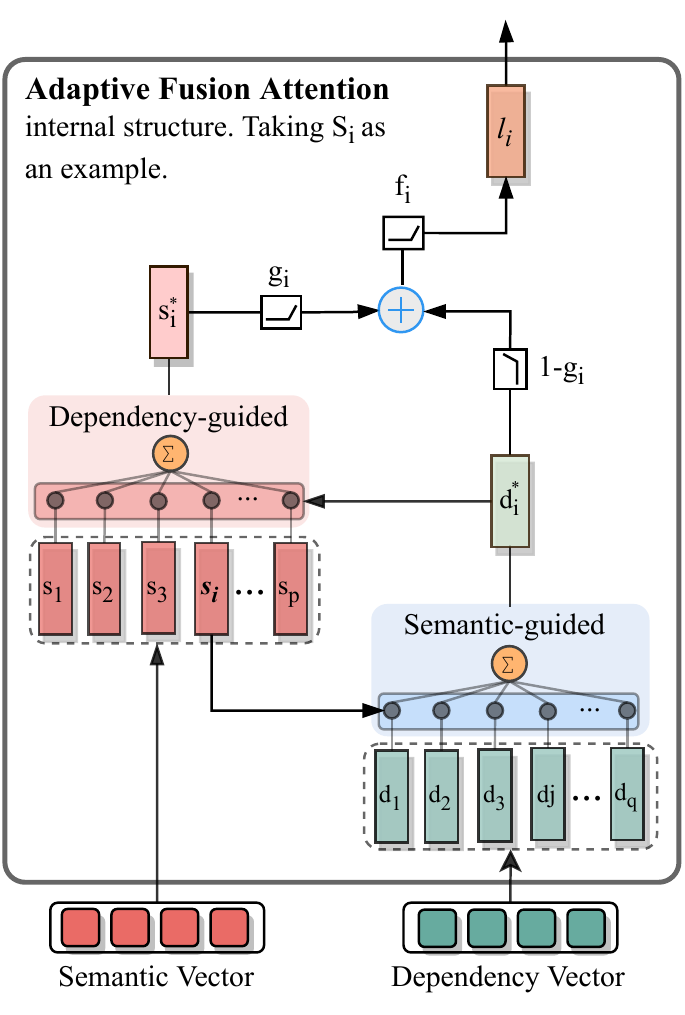}
\caption{\label{fig:Adaptive}The overall structure of Adaptive Fusion Attention.}
\vspace{-0.55cm}
\end{figure}

However, simple concatenation and fusion seem to underestimate the deep interaction between these two signals and ignore the potential noise introduced by dependency structure. Incorrect structural information may produce noisy outputs and give wrong predictions. Therefore, to further improve the fault tolerance rate of our model and reduce the problem of error propagation, we propose an adaptive fusion module. As shown in Figure \ref{fig:Adaptive}, \textbf{\textit{(i)}} it first interacts and aligns two signals flexibly with semantic-guided attention and dependency-guided attention. \textbf{\textit{(ii)}} And then, it adopts multiple gate modules to selectively extract useful features. \textbf{\textit{(iii)}} Finally, due to the possibility of noise, a filtration gate is utilized to adaptively filter out inappropriate information. 

Firstly, we update the dependency signals through semantic-guided attention. We use $\bm{s}_i$ and $\bm{d}_i$ to denote the \textit{i}-th feature of $\mathbf{Sem}$ and $\mathbf{Dep}$ respectively. 
We provide each semantic vector $\bm{s}_i$ to interact with the dependency signals matrix $\mathbf{Dep}$ and obtain the new dependency feature $\bm{d}^*_i$. Then, based on $\bm{d}^*_i$, we can in turn acquire the new semantic feature $\bm{s}^*_i$ through dependency-guided attention. The calculation process is as follows:

{\vspace{-0.3cm} 
\setlength{\abovedisplayskip}{0.1cm}
\setlength{\belowdisplayskip}{0.2cm}
\begin{equation}
\begin{aligned}
\bm{\delta}_i &= tanh(\mathbf{W}_{D} \underline{\mathbf{Dep}} \oplus (\mathbf{W}_{s_i} \bm{s}_i + \bm{b}_{s_i})), \\
\bm{d}^*_i &= \underline{\mathbf{Dep}} * softmax(\mathbf{W}_{d_i} \bm{\delta}_i + \bm{b}_{d_i}), \\
\hspace{-2mm}
\bm{\gamma}_i &= tanh(\mathbf{W}_{S} \underline{\mathbf{Sem}} \oplus (\mathbf{W}_{d^*_i} \bm{d}^*_i+\bm{b}_{d^*_i})), \\
\bm{s}^*_i &= \underline{\mathbf{Sem}} * softmax(\mathbf{W}_{s^*_i} \bm{\gamma}_i + \bm{b}_{s^*_i}),
\end{aligned}
\end{equation}}where ${\mathbf{W}_{D}, \mathbf{W}_{S}, \mathbf{W}_{s_i}, \mathbf{W}_{d^*_i}}\in\mathbb{R}^{d_{seq}*{d_v}}$; $\mathbf{W}_{d_i}$, $\mathbf{W}_{s^*_i}\in\mathbb{R}^{1*2d_{seq}}$; $\bm{b}_{d^*_i}$, $\bm{b}_{s_i}$, $\bm{b}_{d_i}$, $\bm{b}_{s^*_i}$ are weights and bias of our model, and $\oplus$ denotes the concatenation of signal matrix and feature vector. 

Secondly, to adaptively capture and fuse useful information from novel semantic and dependency features, we introduce our gated fusion modules:

{\vspace{-0.3cm} 
\setlength{\abovedisplayskip}{0.1cm}
\setlength{\belowdisplayskip}{0.2cm}
\begin{equation}
\begin{aligned}
\quad \quad \hat{\bm{d}_i} &= tanh(\mathbf{W}_{\hat{d_i}} \bm{d}^*_i + \bm{b}_{\hat{d_i}}), \quad \quad \quad \quad \\
\quad \quad \hat{\bm{s}_i} &= tanh(\mathbf{W}_{\hat{s_i}} \bm{s}^*_i + \bm{b}_{\hat{s_i}}), \quad \quad \quad \quad  \\
\quad g_i &= \sigma(\mathbf{W}_{g_i}(\hat{\bm{d}_i} \oplus  \hat{\bm{s}_i})),  \\
\quad \quad \bm{v}_i &= g_i \hat{\bm{s}_i} + (1 - g_i) \hat{\bm{d}_i}, \quad  \quad \quad \quad
\end{aligned}
\end{equation}}where $\mathbf{W}_{\hat{d_i}}$, $\mathbf{W}_{\hat{s_i}}\in\mathbb{R}^{d_{hid}*{d_v}}$; $\mathbf{W}_{g_i}\in\mathbb{R}^{1*2d_{hid}}$; $\bm{b}_{\hat{d_i}}$, $\bm{b}_{\hat{s_i}}$ are parameters and $d_{hid}$ is the size of hidden layer. $\sigma$ is the sigmoid activation and $g_i$ is the gate that determines the transmission of these two distinct representations. By this way, we get the fusion feature $\bm{v}_i$ that fused the new semantic and dependency signals adaptively.

Eventually, considering the potential noise problem, we propose a filtration gate to selectively leverage the fusion feature. When $\bm{v}_i$ tends to be beneficial, the filtration gate will incorporate the fusion features and the original features. Otherwise, the fusion information will be filtered out:

{\vspace{-0.3cm} 
\setlength{\abovedisplayskip}{0.1cm}
\setlength{\belowdisplayskip}{0.2cm}
\begin{equation}
\begin{aligned}
\quad f_i &= \sigma(\mathbf{W}_{f_i,s_i} (\bm{s}_i \oplus (\mathbf{W}_{v_i} \bm{v}_i + \bm{b}_{v_i} ))), \\
\quad \bm{l}_i &= f_i * tanh(\mathbf{W}_{l_i} \bm{v}_i + \bm{b}_{l_i}),
\end{aligned}
\end{equation}}where $\mathbf{W}_{f_i,s_i}\in\mathbb{R}^{1*2{d_v}}$; $\mathbf{W}_{v_i}$, $\mathbf{W}_{l_i}\in\mathbb{R}^{{d_v}*d_{hid}}$; $\bm{b}_{v_i}$, $\bm{b}_{l_i}$ are trainable parameters and $\bm{l}_i$ is our final dependency-enhanced semantic feature.

\section{Experiments}

The datasets, baselines and all details of our experiments are shown in the Appendix \ref{Details}.

\begin{table*}
\centering
\renewcommand\arraystretch{0.72}
\scalebox{0.8}{
\setlength{\tabcolsep}{2.2mm}{
\begin{tabular}{lcccccccc}
\toprule
\midrule
\multirow{2}*{Model} &\multirow{2}*{Pre-train} & \multicolumn{3}{c}{Sentence Similarity} &\multicolumn{3}{c}{Sentence Inference} &\multirow{2}*{Avg}\\  \cmidrule(r){3-5} \cmidrule(r){6-8}
         ~& &\text{MRPC} & \text{QQP} & \text{STS-B} &\text{MNLI-m/mm}  & \text{QNLI} & \text{RTE}  \\
\midrule
\text{BiMPM$\dagger$\cite{wang2017bilateral}} & \XSolidBrush & 79.6 & 85.0 & - & 72.3/72.1 & 81.4 & 56.4  & - \\
\text{CAFE$\dagger$\cite{tay2017compare}}& \XSolidBrush  & 82.4 & 88.0 & - & 78.7/77.9 & 81.5 & 56.8  & - \\
\text{ESIM$\dagger$\cite{chen2016enhanced}}& \XSolidBrush & 80.3 & 88.2 & - & - & 80.5 & -  & - \\
\text{Transformer$\dagger$\cite{vaswani2017attention}}& \XSolidBrush  & 81.7 & 84.4 & 73.6 & 72.3/71.4 & 80.3 & 58.0  & 74.53 \\
\midrule
\text{BiLSTM+ELMo+Attn$\dagger$\cite{devlin2018bert}} &\Checkmark  & 84.6 & 86.7 & 73.3 & 76.4/76.1 & 79.8 & 56.8  & 76.24 \\
\text{OpenAI GPT$\dagger$} &\Checkmark  & 82.3 & 70.2 & 80.0 & 82.1/81.4 & 87.4 & 56.0  & 77.06 \\
\text{UERBERT$\ddagger$\cite{xia2021using}} &\Checkmark  & 88.3 & 90.5 & 85.1 & 84.2/83.5 & 90.6 & 67.1 & 84.19 \\
\text{SemBERT$\dagger$\cite{zhang2020semantics}} &\Checkmark   & 88.2 & 90.2 & 87.3 & 84.4/84.0 & 90.9 & 69.3  & 84.90 \\
\midrule
\text{BERT-base$\ddagger$\cite{devlin2018bert}}&\Checkmark  & 87.2 & 89.0 & 85.8 & 84.3/83.7 & 90.4 & 66.4  & 83.83 \\
\text{SyntaxBERT-base$\dagger$\cite{bai2021syntax}}&\Checkmark  & \textbf{89.2} & 89.6 & 88.1 & \textbf{84.9}/84.6 & 91.1 & 68.9  & 85.20 \\
\textbf{DAFA-base}$\ddagger$&\Checkmark & 89.0 & \textbf{91.3} & \textbf{89.0} & 84.7/\textbf{84.8} & \textbf{92.3} & \textbf{71.3} & \textbf{86.06} \\
\midrule
\text{BERT-large$\ddagger$\cite{devlin2018bert}}&\Checkmark  & 89.3 & 89.3 & 86.5 & 86.8/85.9 & 92.7 & 70.1  & 85.80 \\
\text{SyntaxBERT-large$\dagger$\cite{bai2021syntax}}&\Checkmark  & \textbf{92.0} & 89.8 & 88.5 & 86.7/86.6 & 92.8 & 74.7  & 87.26 \\
\textbf{DAFA-large}$\ddagger$ &\Checkmark & 91.6 & \textbf{91.8} & \textbf{89.8}& \textbf{87.2}/\textbf{86.9} & \textbf{93.7} & \textbf{76.2} & \textbf{88.17} \\
\midrule
\bottomrule
\end{tabular}}}
\caption{\label{performance}
The performance comparison of DAFA with other methods. We report Accuracy $\times$ 100 on 6 GLUE datasets. Methods with $\dagger$ indicate the results from their paper, while methods with $\ddagger$ indicate our implementation.
}
\vspace{-0.3cm}
\end{table*}

\begin{table}[th]
\centering
\renewcommand\arraystretch{0.9}
\scalebox{0.76}{
\setlength{\tabcolsep}{1.2mm}{
\begin{tabular}{lcccccc}
\toprule
\text{Model} & \text{SNLI} & \text{Sci} & \text{SICK} & \text{Twi} \\
\midrule
\text{ESIM$\dagger$\cite{chen2016enhanced}$\quad$} & 88.0 & 70.6 & - & - \\
\text{CAFE$\dagger$\cite{tay2017compare}$\quad$}  & 88.5 & 83.3 & 72.3 & - \\
\text{CSRAN$\dagger$\cite{tay2018co}$\quad$} & 88.7 & 86.7 & - & 84.0 \\
\midrule
\text{BERT-base$\ddagger$\cite{devlin2018bert}} & 90.7 & 91.8 & 87.2 & 84.8 \\
\text{UERBERT$\ddagger$\cite{xia2021using}} & 90.8 & 92.2 & 87.8 & 86.2 \\
\text{SemBERT$\dagger$\cite{zhang2020semantics}} & 90.9 & 92.5 & 87.9 & 86.8 \\
\text{SyntaxBERT-base$\dagger$\cite{bai2021syntax}} & 91.0 & 92.7 & 88.1 & 87.3  \\
\text{MT-DNN-base$\dagger$\cite{liu2019multi}} & 91.1 & \textbf{94.1} & - & - \\
\textbf{DAFA-base$\ddagger$} & \textbf{91.7} & 93.8 & \textbf{89.8} & \textbf{89.4} \\
\midrule
\text{BERT-large$\ddagger$\cite{devlin2018bert}} & 91.0 & 94.4 & 91.1 & 91.5 \\
\text{SyntaxBERT-large$\dagger$\cite{bai2021syntax}} & 91.3 & 94.7 & 91.4 & 92.1 \\
\text{MT-DNN-large$\dagger$\cite{liu2019multi}} & 91.6 & \textbf{95.0} & - & - \\
\textbf{DAFA-large$\ddagger$} & \textbf{92.1} & 94.8 & \textbf{92.4} & \textbf{92.8} \\
\bottomrule
\end{tabular}}}
\caption{\label{performance4}
The performance on 4 other datasets, including SNLI, Scitail(Sci), SICK and TwitterURL(Twi).}
\vspace{-0.4cm}
\end{table}

\subsection{Results}

In our experiments, we implemented DAFA in the initial transformer layer of BERT.

\begin{table*}[t]
\centering
\setcounter{table}{6}
\renewcommand\arraystretch{0.95}
\scalebox{0.82}{
\setlength{\tabcolsep}{2mm}{
\begin{tabular}{lcccc}
\toprule
\midrule
\text{Case} & \text{BERT} & \text{DAFA-avg} & \text{DAFA} & \text{Gold} \\
\midrule
\text{S1:}Please help me book a flight \textcolor{red}{from New York to Seattle}. & \multirow{2}{*}{\text{label:1}} & \multirow{2}{*}{\text{label:0}} & filter gate:0.93 & \multirow{2}{*}{\text{label:0}} \\
\text{S2:}Please help me book a flight \textcolor{blue}{from Seattle to New York}. & ~ & ~ & label:0 & ~ \\
\midrule
\text{S1:}How does \textcolor{red}{reading help you think better}? & \multirow{2}{*}{\text{label:1}} & \multirow{2}{*}{\text{label:0}} & filter gate:0.91 & \multirow{2}{*}{\text{label:0}} \\
\text{S2:}How do \textcolor{blue}{you think it's better to read}? & ~ & ~ & label:0 & ~ \\
\midrule
\text{S1:}Sorry, \textcolor{red}{I got sick yesterday} and couldn't \textcolor{red}{have lunch with you}. & \multirow{2}{*}{\text{label:1}} & \multirow{2}{*}{\text{label:0}} & filter gate:0.15 & \multirow{2}{*}{\text{label:1}} \\
\text{S2:}Sorry, \textcolor{blue}{I was taken ill yesterday} and unable to \textcolor{blue}{meet you for lunch}. & ~ & ~ & label:1 & ~ \\
\midrule
\text{S1:}\textcolor{red}{The largest lake in America} is in my hometown \textcolor{red}{called Lake Superior}. & \multirow{2}{*}{\text{label:1}} & \multirow{2}{*}{\text{label:0}} & filter gate:0.11 & \multirow{2}{*}{\text{label:1}} \\
\text{S2:}\textcolor{blue}{Lake Superior is the largest lake in America}, \textcolor{blue}{and it's} in my hometown. & ~ & ~ & label:1 & ~ \\
\midrule
\bottomrule
\end{tabular}}}
\caption{\label{example} The example sentence pairs of our cases. \textcolor{red}{Red} and \textcolor{blue}{Blue} are difference phrases in sentence pair. 
DAFA-avg means replacing the adaptive fusion module with a simple average of semantic and the dependency signals.
}
\vspace{-0.3cm}
\end{table*}

\begin{table}
\centering
\renewcommand\arraystretch{0.8}
\scalebox{0.83}{
\setlength{\tabcolsep}{3.5mm}{
\begin{tabular}{lc}
\toprule
  \multirow{2}{*}{\text{Method}} & \text{QQP} \\
  \cmidrule(lr){2-2}
  ~ & \text{DEV \quad Test} \\
\midrule
\textbf{DAFA$\ddagger$} & \textbf{92.5 \quad 91.8} \\
\midrule
\; \text{w/o \ \ dependency tree simility$\ddagger$} & 91.7 \quad 91.1 \\
\; \text{w/o \ \ subgraph matching$\ddagger$} & 92.0 \quad 91.3 \\
\; \text{w/o \ \ tf-idf weights$\ddagger$} & 92.2 \quad 91.6 \\
\midrule
\; \text{w/o \ \ cross attention$\ddagger$} & 91.8 \quad 91.2 \\
\; \text{w/o \ \ gate mechanism$\ddagger$} & 91.9 \quad 91.4 \\
\; \text{w/o \ \ adaptive fusion+averaging$\ddagger$} & 91.5 \quad 91.0 \\
\bottomrule
\end{tabular}}}
\caption{\label{ablation}
The ablation experiment results of our method.
}
\vspace{-0.4cm}
\end{table}

\textbf{The Main Results of GLUE Datasets} We first fine-tune our model on 6 GLUE datasets. Table \ref{performance} shows the performance of DAFA and other competitive models. It can be seen that using only \textit{non-pretrained models} performs obviously worse than PLMs due to their strong context awareness and data fitting capabilities.
When the backbone model is BERT-base and BERT-large, the average accuracy of DAFA respectively improves by \textbf{2.23\%} and \textbf{2.37\%} than vanilla BERT. Such great improvement demonstrates the benefit of adaptive fusion dependency structure for mining semantics, and also proves that our framework can help BERT to perform much better in SSM.

Moreover, some previous works such as SemBERT, UERBERT and SyntaxBERT also outperform vanilla BERT by injecting external knowledge, but DAFA still maintains the best performance. 
Specifically, our model outperforms SyntaxBERT, the top-performing model in previous work leveraging external knowledge, with an average relative improvement of \textbf{0.91\%} based on BERT-large. Especially on the QQP dataset, the accuracy of DAFA is significantly improved by \textbf{2\%} over SyntaxBERT. There are two main reasons: 
\begin{itemize}[itemsep=0pt, topsep=2pt, leftmargin=10pt]
\setlength{\itemindent}{0pt}
\setlength{\parskip}{0pt}
\item[$\diamond$] On the one hand, we use subgraph matching and keyword weights to enhance the ability of DAFA to capture dependency knowledge. DAFA obtains interactive information that is more conducive to fusing fine-grained features.
\item[$\diamond$] On the other hand, for the latent noise introduced by external knowledge, our adaptive fusion module can selectively filter out inappropriate signals to suppress the propagation of noise, while previous work seems to have not paid enough attention to this issue. However, we still notice that SyntaxBERT achieves slightly better accuracy on few datasets. We consider this to be a result of the instability of noise.
\end{itemize}

\textbf{The Results of Other Datasets} Second, to verify the general performance of our method, we also conduct experiments on 4 other popular datasets. The results are shown in Table \ref{performance4}. DAFA still outperforms vanilla BERT and some representative models on almost all datasets. It is worth noting that although DAFA outperforms MT-DNN \cite{liu2019multi} on SNLI, it does not perform as well as MT-DNN on Scitail. 
This is because MT-DNN \cite{liu2019multi} uses more model parameters and a large amount of cross-tasks training data, which makes MT-DNN more advantageous in this regard. But MT-DNN will also require more training time and computational cost. Besides, the data volume of Scitail is relatively small, which makes the \textit{variance} of the model prediction results larger.
However, DAFA still shows a very competitive performance on Scitail, which also shows from the side that our method can make up for the lack of generalization ability with fewer parameters by endowing BERT with dependency structure awareness.

\textbf{Overall}, consistent conclusions can be drawn from such results. Compared to previous works, our \textit{dependency framework} is highly competitive in further judging semantic similarity, and the experimental results also confirm our thoughts. 

\subsection{Ablation Study}

To assess the contribution of each component in our approach, we have ablation experiments on the QQP dataset based on BERT-large. The experiment results are shown in Table \ref{ablation}.

\textbf{Dependency Matrix} Our \textit{dependency matrix} is jointly constructed by three components:
\textbf{\textit{(a)}} We first remove the dependency tree similarity and subgraph matching module respectively, and the model performance dropped by nearly \textbf{0.7\%} and \textbf{0.5\%}. This suggests that simple dependency structure alignment can further describe the interactions between words to achieve better semantic similarity. \textbf{\textit{(b)}} Then, subgraph matching can align the dependency substructures and enrich the contextual representation by introducing finer-grained comparison information. \textbf{\textit{(c)}} Besides, due to the different importance of words in the sentence, \textit{tf-idf} can weight each word to readjust the attention distribution. The accuracy also dropped slightly after the model removed \textit{tf-idf}. The above experiments demonstrate the effectiveness of each component of our \textit{dependency matrix}. 

\textbf{Adaptive Fusion} We also conduct multiple experiments to verify the effect of adaptively fusing the original semantic signals and the dependency signals. \textbf{\textit{(a)}} We first remove the \textit{cross-attention module}, and the performance drops to \textbf{91.2\%}. Cross-attention can capture the interaction information between two signals, and interactivity information is crucial for semantic matching. \textbf{\textit{(b)}} Moreover, we remove \textit{multiple gate mechanisms}, only relying on the attention modules to integrate our signals. And the accuracy drops to \textbf{91.4\%}. It shows that the ability of the model to suppress noise is weakened without filter gates.
We also replace the overall adaptive fusion module with \textit{simply averaging} and the performance drops sharply to \textbf{91.0\%}, indicating that soft aggregation and governance can better integrate semantic and dependency signals.

\begin{figure}
\centering
\includegraphics[width=0.48\textwidth]{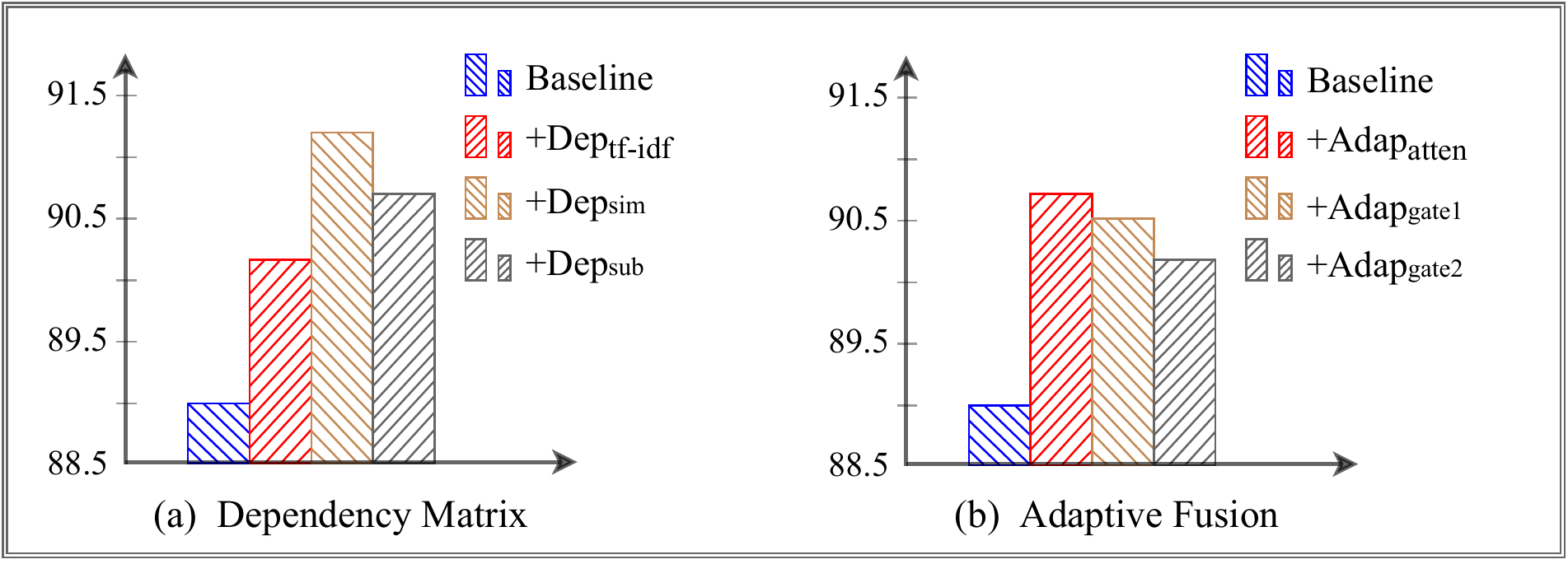}
\caption{\label{fig:influence}Effects of different sub-module integration methods on the QQP test set. \textit{$Dep_{sim}$}, \textit{$Dep_{sub}$}, \textit{$Dep_\textit{tf-idf}$} indicate only used sample dependency, subgraphs matching, \textit{tf-idf} weights in dependency matrix respectively. \textit{$Adap_{atten}$}, \textit{$Adap_{gate1}$}, \textit{$Adap_{gate2}$} represent models that only assemble cross attention, the first gate and the second gate in adaptive fusion respectively.}
\vspace{-0.3cm}
\end{figure}

\textbf{Sub-Module Analysis} To further verify the contribution of each submodule to DAFA, we separately assemble the respective sub-modules in dependency matrix computation and adaptive fusion, the results are shown in Figure \ref{fig:influence}. First, we can find that after adding any of sub-modules, the performance of the model is improved over the baseline. Second, the aggregation of dependency similarity and fusion attention provides the most significant improvements, which intuitively reflects their cornerstone roles as core modules of dependency matrix and adaptive fusion respectively. Such results confirm the necessity of each sub-module again.

\textbf{Overall}, due to the efficient combination of each component, DAFA can adaptively fuse dependency aligned features into pre-trained models and leverage their powerful contextual representation capability to better inference semantics.

\subsection{Case Study}

To visually demonstrate the validity of our approach, we also conduct a qualitative study using multiple cases in Table \ref{example}. 

S1 and S2 in the first two cases are literally similar and differ only in the dependency between words, but they express two quite different semantics. In the first example, BERT attempts to capture interaction information from these two sentences, but ignores the dependency between \textit{“New York”} and \textit{“Seattle”}. It fails to distinguish the semantic difference and gives wrong prediction results.
By adopting the dependency structure, our method can capture dependency dislocation information and comprehend fine-grained semantics. As the results show, DAFA gives correct predictions.

However, the injection of dependency structure may generate noise and interference. 
For example, in the third case, \textit{“got sick”} and \textit{“was token ill”} express the same semantics, but their dependency trees diverge significantly and may mislead the model. By simply averaging the semantic and dependency signals, DAFA-avg instead gives the wrong answers in the last two examples. 

Therefore, we propose an adaptive fusion module to reduce the possibility of the noise or useless signals. The filter gate reflects the degree to which the model adopts the dependency structure. In the first two examples, our model learned the important impact of dependency on semantics by adaptively fusing distinct information. DAFA automatically sets the filter gate to 0.91-0.93 and improves perception of dependency structure. However, in the last two examples, our adaptive fusion module identifies the latent noise in dependency structure. To alleviate the possible misleading effect, DAFA correspondingly sets the filter gate to about 0.1, which weakens the model's sensitivity to dependency.

Eventually, as the results show, our DAFA makes correct predictions in all of the above cases and increases the fault tolerance of the model. Such results again demonstrate the effectiveness and necessity of our components. 

\section{Qualitative Analysis}

\begin{figure}
\centering
\includegraphics[width=0.44\textwidth]{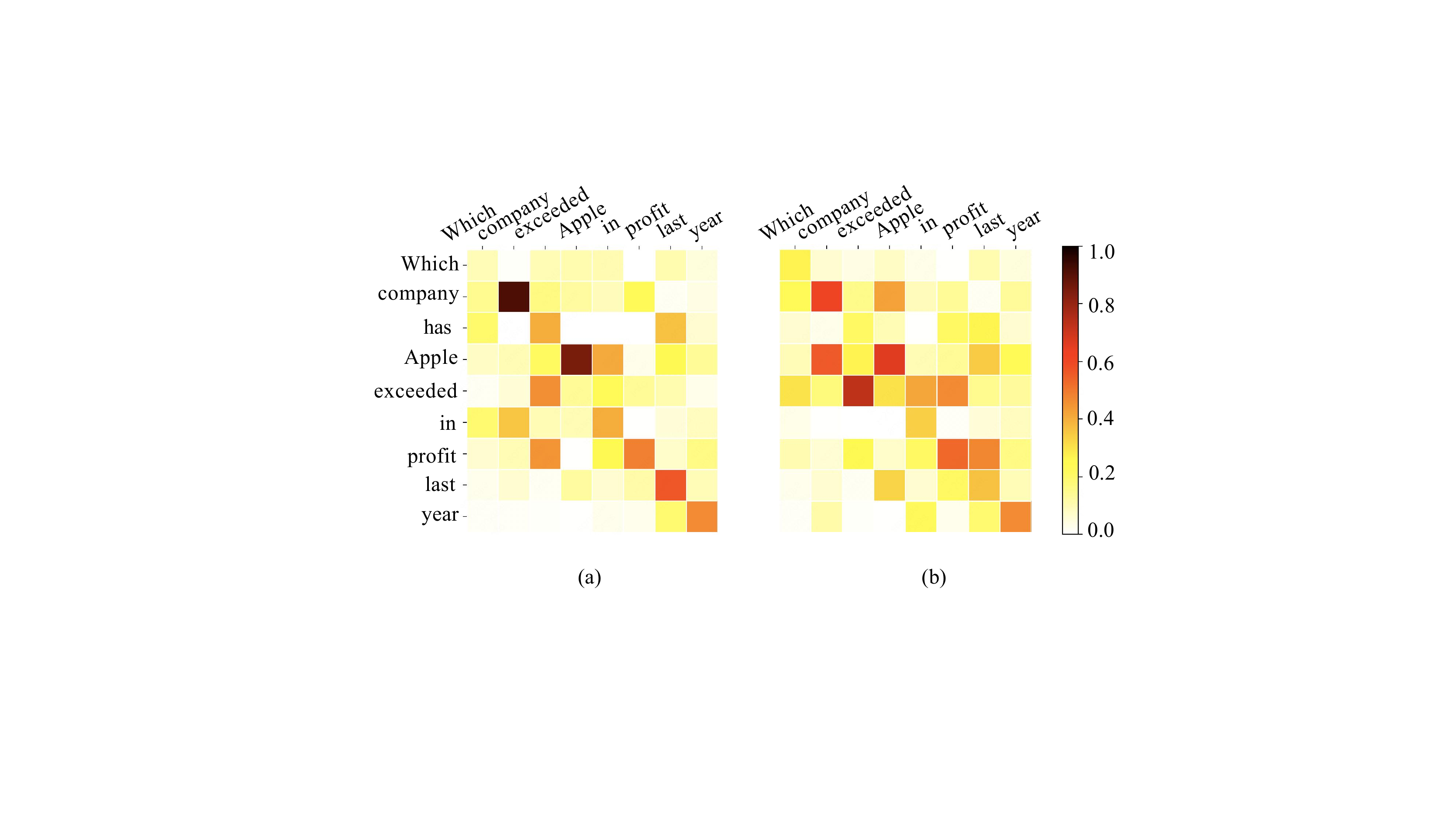}
\caption{\label{fig:weight}The attention weight distribution of BERT (a) and our method (b).}
\vspace{-0.55cm}
\end{figure}

\textbf{Attention Distribution and Interpretability} To verify the calibration effect of the dependency structure alignment as well as to perform a visual interpretability analysis, we display the attention distribution between two sentences that are literally similar but semantically distinct. The weights for one of our attention heads are shown in Figure \ref{fig:weight}.

Obviously, vanilla BERT is heavily influenced by the same words in sentences. It ignores deep semantic associations and instead over-focuses on shallow literal features, which may lead to wrong predictions. 
However, after being calibrated by our method, the attention weights not only learn the shared word in sentences, but also pay more attention to the alignment between the dependency structures. For example, DAFA not only increases the weight of the same \textit{“exceeded”} in the two sentences, but also increases the weight between \textit{“Apple”} in the first sentence and \textit{“company”} in the second one. This is because \textit{“Apple-exceeded”} and \textit{“company-exceeded”} are subject-predicate dependency structures in the two sentences respectively.

Meanwhile, attention modules are often used to explore the interpretability of the model \cite{clark2019does,hao2020self,lin2019open}. 
In Figure \ref{fig:weight}(a), we can observe that it is difficult to determine the deep interaction information between sentences by simple contextual features. However, dependency can align sentence structure at the word level. In Figure \ref{fig:weight}(b), DAFA significantly mitigates the strong influence of the same word and enhances the sensitivity to dependency structures. The calibrated attention distribution is more in line with human cognition and validates our methodology. Since dependency and semantics are linguistic expressions with different perspectives and granularities, their combination can further improve the model's awareness to discern subtle semantic differences and reduce error propagation problems. In addition, the results of ablation experiments also confirm DAFA at each component level and provide results consistent with our predictions. 
Therefore, our method provides better interpretability.

\begin{figure}
\centering
\includegraphics[width=0.45\textwidth]{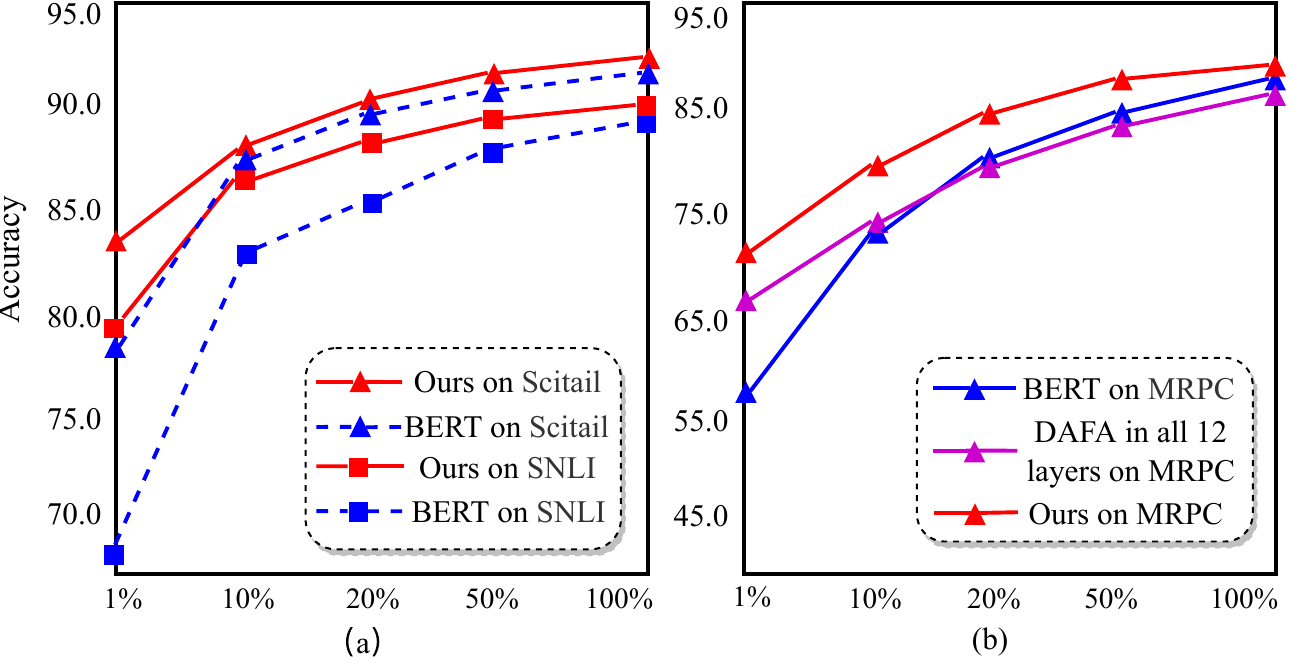}
\caption{\label{fig:curve}(a) The performance of our model and BERT in different data scenarios of SNLI and Scitail. (b) The performance of implementing DAFA in different layer of BERT on MRPC. The X-axis: the amount of data.}
\vspace{-0.5cm}
\end{figure}

\textbf{Data Scenarios and Structural Analysis} To further verify the generalization ability of our method, we conduct a range of experiments under different data scenarios on SNLI and Scitail. We use 1\% to 100\% of the train set corpus to fine-tune our model, and then examine it on the test set. As illustrated in Figure \ref{fig:curve}(a), our approach obviously improves the performance of BERT in data sparsity scenarios (1\%) and always surpasses BERT at different amounts of data (from 1\% to 100\%). This shows that dependency prior knowledge provides highly salient performance gain happens when the train data is few, and further proves that DAFA can effectively enhance BERT on distinct data scenarios.

To explore which layer most requires the dependency structure, we implement DAFA in the initial transformer layer and in all 12 transformer layers of BERT respectively. Our experiments use 1\% to 100\% data of MRPC and the results are shown in Figure \ref{fig:curve}(b). The effect of our method (initial layer) significantly exceeds vanilla BERT and the approach that implements DAFA in all 12 layers of BERT. The main reason is that BERT pays more attention to word-level features at the bottom layers and semantic features at the top layers.

\begin{figure}
\centering
\includegraphics[width=0.45\textwidth]{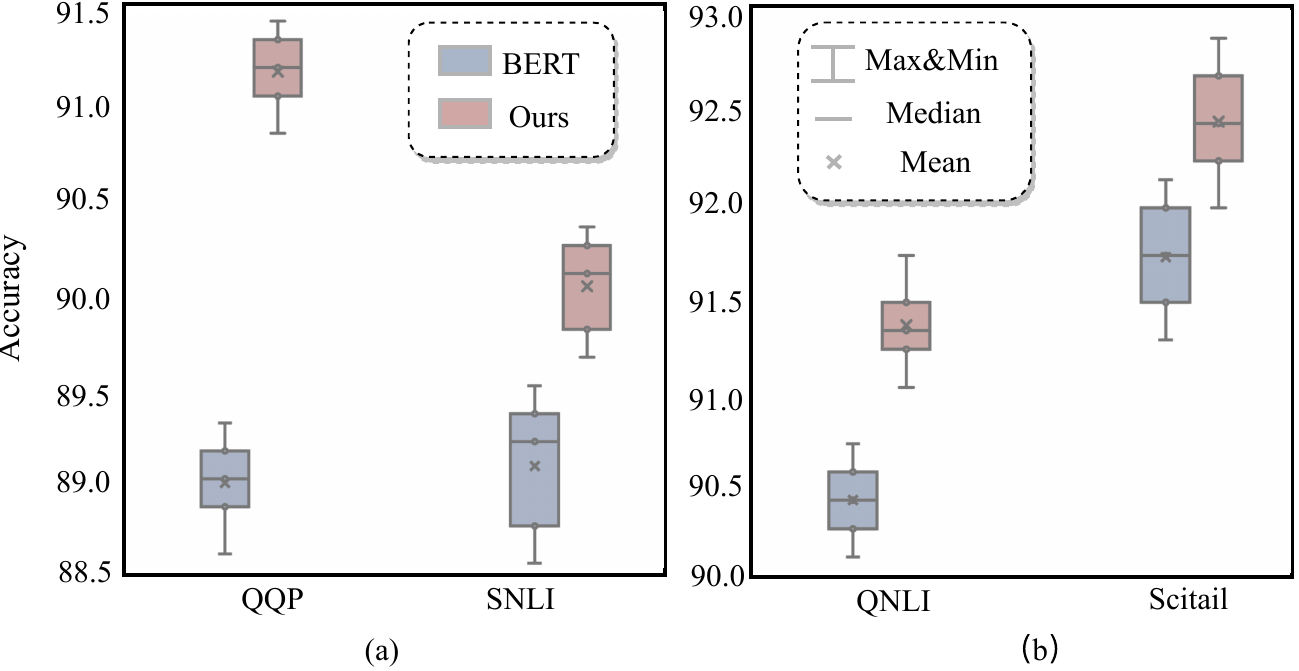}
\caption{\label{fig:stability}The stability of DAFA and BERT on 4 datasets (QQP, SNLI, QNLI, Scitail).}
\vspace{-0.53cm}
\end{figure}

\textbf{Stability Study} We also conducted extensive experiments on 4 datasets for exploring the stability of our method. To minimize the effect of randomness in BERT training, performance levels are averaged across 10 different runs on the dev set. The performance distribution box plot is shown in Figure \ref{fig:stability}. The median and mean levels of our model surpass vanilla BERT on all 4 datasets, and the performance fluctuation range of our method is within ±1\% around the mean levels, which indicates that our method has better stability relative to BERT on different data distributions.

\section{Related Work}

\textbf{Semantic Sentence Matching} is a fundamental task in NLP. In recent years, thanks to the appearance of large-scale annotated datasets \cite{bowman2015large,williams2017broad}, neural network models have made great progress in SSM \cite{qiu2015convolutional,wan2016deep}, mainly fell into two categories. The first one \cite{conneau2017supervised,choi2018learning} focuses on encoding sentences into corresponding vector representations without any cross-interaction and applies a classifier layer to obtain similarity. The second one \cite{liang2019asynchronous,chen2016enhanced,xue2023dual} utilizes cross-features as an attention module to express the word-level or phrase-level alignments, and aggregates these integrated information to acquire similarity. 

Recently, the shift from neural network architecture engineering to large-scale pre-training has achieved outstanding performance in SSM and many other tasks. Meanwhile, leveraging external knowledge \cite{miller1995wordnet,bodenreider2004unified} to enhance PLMs has been proven to be highly useful for multiple NLP tasks \cite{kiperwasser2018scheduled}. Therefore, recent work attempts to integrate external knowledge into pre-trained language models, such as AMAN, DABERT, UERBERT, SyntaxBERT, and so on \cite{liang2019adaptive,wang-etal-2022-dabert,xia2021using,bai2021syntax}.

\textbf{Dependency Syntax} As a crucial prior knowledge, dependency tree provides a form that is able to indicate the existence and type of linguistic dependency relation among words, which has been shown general benefits in various NLP tasks \cite{bowman2016fast}. Therefore, many approaches that adopted syntactic dependency information have been proposed and attained multiple great results \cite{duan2019syntax}. For example, \citet{strubell2018linguistically} present a linguistically-informed self-attention (LISA) in a multi-task learning framework. \citet{sachan2020syntax} investigate popular strategies for incorporating dependency structure into PLMs. \citet{liu2023time} used a grammar-guided dual-context architecture network (SG-Net) to achieve SOTA effects on span-based answer extraction tasks.

\section{Conclusions}
\label{sec:bibtex}
In this paper, we propose a Dependency-Enhanced Adaptive Fusion Attention (DAFA), which can adaptively utilize dependency alignment features for semantic matching. Based on the context representation capability of BERT, DAFA enables the model to learn more fine-grained comparison information and enhances the sensitivity of PLMs to the dependency structure. The experiment results on 10 public datasets indicate that our approach can achieve better performance than multiple strong baselines. Since DAFA is an end-to-end trained component, it is expected to be applied to other large-scale pre-trained models in the future.


\bibliography{custom}
\bibliographystyle{acl_natbib}

\clearpage
\appendix
\section{Appendix}
\label{sec:appendix}
\renewcommand{\thefootnote}{\arabic{footnote}}
\begin{figure*}[ht]
\centering
\includegraphics[width=1.0\textwidth]{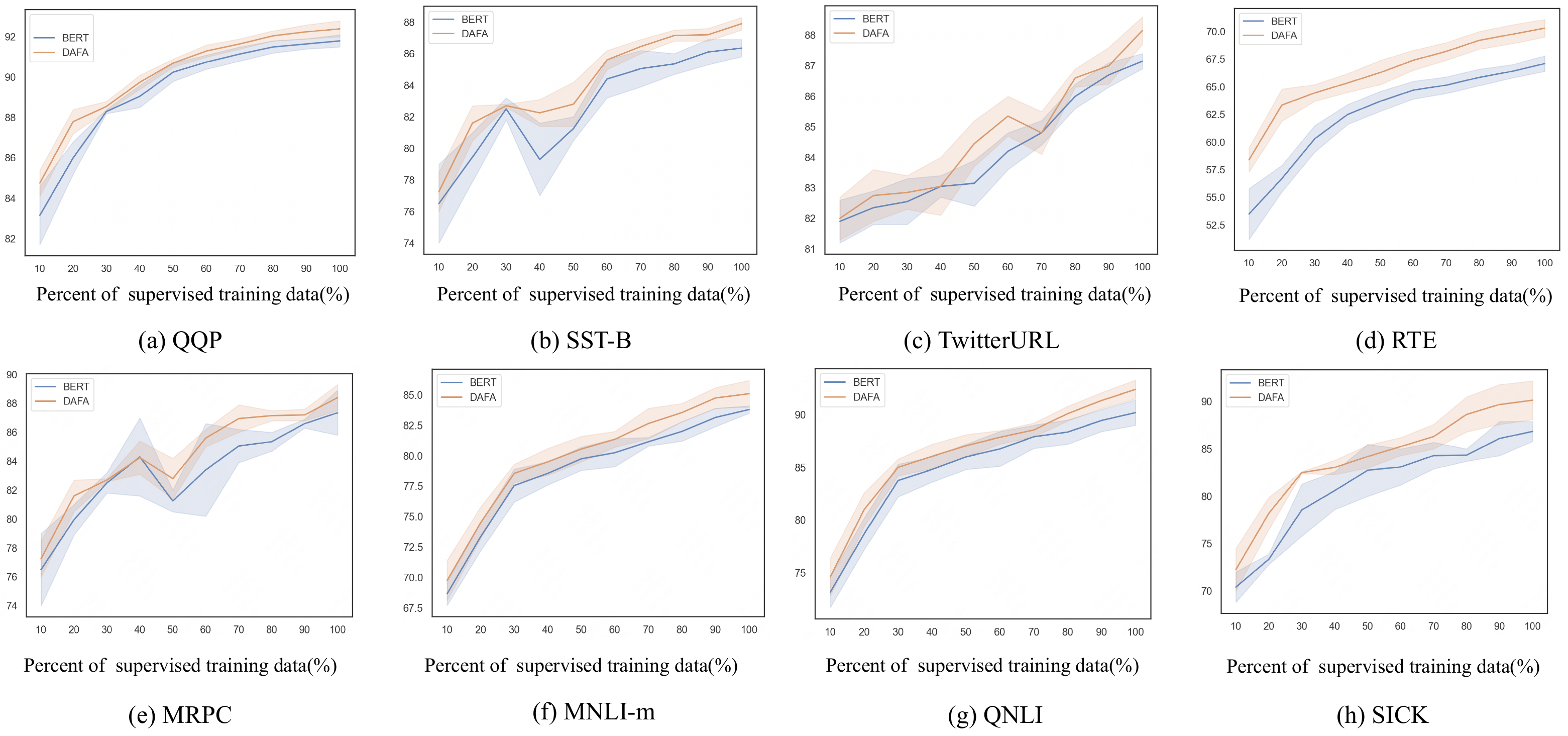}
\caption{\label{fig:performance} Performance of BERT and DAFA with different amounts of training data. X-axis: Percent of supervised training data. Y-axis: Accuracy of model. The colored bands indicate ±1 standard deviation corresponding to different percentages of training data.}
\vspace{-0.4cm}
\end{figure*}

\subsection{Effect of Different Training Data Volumes}
\label{DataVolumes}

We randomly select 10\% to 100\% of the data from the training data and conduct data scene analysis experiments on eight other datasets. We show the results in Figure \ref{fig:performance}. For BERT and DAFA, we have trained 5 times for each training scale of each dataset. The changing curves reveal many interesting patterns. First, the performance of our proposed method outperforms vanilla BERT almost uniformly across all training data sizes. Second, on datasets such as RTE, STS-B, and SICK, dependency provide the most significant performance improvement when the training data is small. These findings suggest that if training data is scarce, it is wise to consider injecting dependency knowledge into BERT.

\section{Appendix}
\label{Details}
\subsection{Implementation Details of Our Experiments}

\textbf{Implementation Details} DAFA is based on BERT-base and BERT-large. For distinct targets, our hyper-parameters are different. We use the AdamW optimizer and set the learning rate in \{$1e^-5$, $2e^-5$, $3e^-5$, $8e^-6$\}. We set warm-up 0.1, $L2$ weight decay $1e^-8$ and constant $\theta$ is 2.0. Our epoch is between 3 and 5, and the batch size is selected in \{16, 32, 64\}. We also set dropout at 0.1-0.3 and gradient clipping in \{7.5, 10.0, 15.0\}. Our experiments are performed one A100 GPU.
Besides, our dependency parser is the biaffine parser proposed by \citet{dozat2016deep}. We use original phrase-structure Penn Treebank (PTB) \cite{marcinkiewicz1994building} to convert by the Stanford Parser v3.3.02\footnote{https://nlp.stanford.edu/software/lex-parser.shtml} to retrain a parser model. The dependency parser is not updated with our framework.

\subsection{Datasets Statistics}
\label{Datasets}

The statistics of all 10 datasets is shown in Table \ref{statistics}.

\subsubsection{GLUE Datasets}

We experimented with 6 datasets of the GLUE\footnote{https://huggingface.co/datasets/glue} datasets \cite{wang2018glue}: MRPC, QQP, STS-B, MNLI-m/mm, QNLI and RTE, The following is a detailed introduction to these datasets.

\begin{itemize}[itemsep=0pt, topsep=2pt, leftmargin=10pt]
\setlength{\itemindent}{0pt}
\setlength{\parskip}{0pt}
\item[$\circ$] \textbf{MRPC} is a dataset that automatically extract sentence pairs from online news sources and manually annotate whether the sentences in sentence pairs are semantically equivalent. The task is to determine whether there are two categories of interpretation: interpretation or not interpretation.
\item[$\circ$] \textbf{QQP} comes from the famous community Q\&A website quora. Its goal is to predict which of the provided question pairs contains two questions with the same meaning.
\item[$\circ$] \textbf{STS-B} is a collection of sentence pairs extracted from news headlines, video titles, image titles and natural language inference data. Each pair is annotated by humans, and its similarity score is 0-5. The task is to predict these similarity scores, which is essentially a regression problem, but it can still be classified into five text classification tasks of sentence pairs.
\item[$\circ$] \textbf{MNLI-m/mm} is a crowd-sourced collection of sentence pairs annotated with textual entailment information. Given the promise statement and hypothesis statement, the task is to predict whether the premise statement contains assumptions (entailment), conflicts with assumptions (contradiction), or neither (neutral). 
\item[$\circ$] \textbf{QNLI} is a question and answer data set composed of a question paragraph pair, in which the paragraph is from Wikipedia, and a sentence in the paragraph contains the answer to the question. The task is to judge whether the question and sentence (sentence, a sentence in a Wikipedia paragraph) contain, contain and do not contain, and classify them. 
\item[$\circ$] \textbf{RTE} is a series of datasets from the annual text implication challenge. These data samples are constructed from news and Wikipedia. All these data are converted into two categories. For the data of three categories, neutral and contradiction are converted into not implication in order to maintain consistency. 
\end{itemize}

\subsubsection{Other Datasets}

We also experimented with 4 other popular datasets :SNLI\footnote{https://nlp.stanford.edu/projects/snli/}, Scitail\footnote{https://allenai.org/data/scitail}, SICK\footnote{http://marcobaroni.org/composes/sick.html} and TwitterURL\footnote{https://github.com/lanwuwei/Twitter-URL-Corpus}.
The following is an introduction to these 4 datasets.
\begin{itemize}[itemsep=0pt, topsep=2pt, leftmargin=10pt]
\setlength{\itemindent}{0pt}
\setlength{\parskip}{0pt}
\item[$\circ$] \textbf{SNLI}\cite{bowman2015large} is a popular dataset used for entailment classification (or natural language inference). The task is to determine whether two sequences entail, contradict or are mutually neutral. 
\item[$\circ$] \textbf{Scitail}\cite{khot2018scitail} is an entailment dataset created from multiple-choice science exams and web sentences. Each question and the correct answer choice are converted into an assertive statement to form the hypothesis. 
\item[$\circ$] \textbf{SICK}\cite{marelli2014sick} is a dataset for semantic textual similarity estimation. The task is to assign a similarity score to each sentence pair.
\item[$\circ$] \textbf{TwitterURL}\cite{lan2017continuously} is a collection of sentence level paraphrases from Twitter by linking tweets through shared URLs. Its goal is to discriminate duplicates or not.
\end{itemize}

\begin{table}
\centering
\renewcommand\arraystretch{0.9}
\setlength{\tabcolsep}{2.2mm}{
\scalebox{0.78}{
\begin{tabular}{lccccc}
\toprule
\text{Datasets} & \text{\#Train} & \text{\#Dev} & \text{\#Test} & \text{\#Class} \\
\midrule
\text{MRPC} & 3669 & 409 & 1380 & 2 \\
\text{QQP} & 363871 & 1501 & 390965 & 2 \\
\text{MNLI-m/mm} & 392703 & 9816/9833 & 9797/9848 & 3 \\
\text{QNLI} & 104744 & 40432 & 5464 & 2 \\
\text{RTE} & 2491 & 5462 & 3001 & 2 \\
\text{STS-B} & 5749 & 1500 & 1379 & 2 \\
\text{SNLI} & 549367 & 9842 & 9824 & 3 \\
\text{SICK} & 4439 & 495 & 4906 & 3 \\
\text{Scitail} & 23596 & 1304 & 2126 & 2 \\
\text{TwitterURL} & 42200 & 3000 & 9324 & 2 \\
\bottomrule
\end{tabular}}}
\caption{\label{statistics}
The statistics of all 10 datasets.
}
\vspace{-0.4cm}
\end{table}

\subsection{Baselines}
\label{Baselines}

To evaluate the effectiveness of our proposed DAFA in SSM, we mainly introduce BERT \cite{devlin2018bert}, SemBERT \cite{zhang2020semantics}, SyntaxBERT, UERBERT \cite{xia2021using} and multiple PLMs \cite{devlin2018bert} for comparison. Moreover, we also selected several competitive no pre-trained models as baselines, such as ESIM \cite{chen2016enhanced}, Transformer \cite{vaswani2017attention} , etc \cite{hochreiter1997long,wang2017bilateral,tay2017compare}.
\begin{itemize}[itemsep=0pt, topsep=2pt, leftmargin=10pt]
\setlength{\itemindent}{0pt}
\setlength{\parskip}{0pt} 
\item \textbf{BIMPM} is propsed in \cite{wang2017bilateral} and employs a multi-perspective matching mechanism in sentence pair modeling tasks.
\item \textbf{CAFE} \cite{tay2017compare} introduces a new architecture where alignment pairs are compared, compressed and then propagated to upper layers for enhanced representation learning. And then it adopts factorization layers for efficient and expressive compression of alignment vectors into scalar features, which are then used to augment the base word representations. 
\item \textbf{ESIM} \cite{chen2016enhanced} proved that the sequential inference model based on chained LSSM can outperform previous complex structures. It futher achieved new SOTA performances.
\item \textbf{CSRAN} \cite{tay2018co} is a deep architecture, involving stacked recurrent encoders. CSRAN incorporates two novel components to take advantage of the stacked architecture. It first introduces a new bidirectional alignment mechanism that learns affinity weights by fusing sequence pairs across stacked hierarchies. And then it leverages a multi-level attention refinement component between stacked recurrent layers.
\item \textbf{Transformer} \cite{vaswani2017attention} uses the attention mechanism to reduce the distance between any two positions in the sequence to a constant. It is not a sequential structure similar to RNN, so it has better parallelism.
\item \textbf{ELMO} \cite{peters2018deep} adopts a typical two-stage process. The first stage is pre training using language model; The second stage is to extract the word embedding of each layer of the network corresponding to the word from the pre training network and add it to the downstream task as a new feature. It can solve the problem of polysemy of the previous language model, because the generated word vector is changed according to the change of the specific use context.
\item \textbf{GPT} is a semi-supervised learning method that uses a large amount of unlabeled data to let the model learn "common sense" to alleviate the problem of insufficient labeled information. The specific method is to pre-train the model Pretrain with unlabeled data before training Fine-tune for labeled data, and ensure that the two kinds of training have the same network structure.
\item \textbf{BERT} \cite{devlin2018bert} Given that our model implements based on BERT, we naturally compare it with vanilla BERT without prior knowledge. We adopt the configuration of Google’s BERT-base in our experiments. 
\item \textbf{UERBERT} \cite{xia2021using} conducted lots of experiments to analyze which kind of external knowledge that BERT has already known, and directly injected the synonym knowledge into BERT without fine-tuning.
\item \textbf{SemBERT}\cite{zhang2020semantics} incorporates explicit contextual semantics from pre-trained semantic role labeling and is capable of explicitly absorbing contextual semantics over a BERT backbone. SemBERT keeps the convenient usability of its BERT precursor in a light fine-tuning way without substantial task-specific modifications.
\item \textbf{Syntax-BERT}  is a framework that integrate the syntax trees into transformer-based models. Unlike us, it explicitly injected syntactic knowledge into checkpoints of models.
\item \textbf{MT-DNN}\cite{liu2019multi} not only leverages large amounts of cross-task data, but also benefits from a regularization effect that leads to more general representations to help adapt to new tasks and domains. MT-DNN extends the model by incorporating a pre-trained bidirectional transformer language model.
\end{itemize}

\end{document}